\documentclass[10pt,journal,compsoc]{IEEEtran}
\usepackage{graphicx}
\usepackage{ragged2e}
\usepackage{amsmath}
\usepackage{caption}
\usepackage{amssymb}
\usepackage{longtable,booktabs}
\usepackage{fancyhdr}
\usepackage{multirow}
\usepackage{footmisc}
\usepackage[comma,numbers,square,sort&compress]{natbib}

\begin{document}
\title{Predictive Learning: Using Future Representation Learning Variantial Autoencoder for Human Action Prediction}
\author{Yu Runsheng$^1$, Shi Zhenyu$^1$,Ma Qiongxiong$^1$, Qing Laiyun$^{2}$\\
$^1$ South China Normal University\\
$^2$ University of Chinese Academy of Sciences\\
$\{$20143201002,20143201094$\}$@m.scnu.edu.cn
}

\IEEEtitleabstractindextext{
\justifying   

\begin{abstract}
The unsupervised Pretraining method has been widely used in aiding human action recognition. However, existing methods focus on reconstructing the already present frames rather than generating frames which happen in future.In this paper, We propose an improved Variantial Autoencoder model to extract the features with a high connection to the coming scenarios, also known as Predictive Learning. Our framework lists as following: two steam 3D-convolution neural networks are used to extract both spatial and temporal information as latent variables. Then a resample method is introduced to create new normal distribution probabilistic latent variables and finally, the deconvolution neural network will use these latent variables generate next frames. Through this possess, we train the model to focus more on how to generate the future and thus it will extract the future high connected features. \\
In the experiment stage, A large number of experiments on UT and UCF101 datasets reveal that future generation aids Prediction does improve the performance. Moreover, the Future Representation Learning Network reach a higher score than other methods when in half observation. This means that Future Representation Learning is better than the traditional Representation Learning and other state- of-the-art methods in solving the human action prediction problems to some extends.

\end{abstract}

}

\maketitle
\IEEEdisplaynontitleabstractindextext
\IEEEpeerreviewmaketitle
\IEEEraisesectionheading{\section{Introduction}\label{sec:introduction}}
\IEEEPARstart{A}{ction} prediction, also known as early event recognition, aims to recognize an action before it happens.It plays a more and more significant role in both life and industry.For example, it can stop someone commits crimes automatically before it occurs from the surveillance system as well as respond correctly according to human action in robot-human interaction field.
\begin{figure}[!htb]
\centering
{\includegraphics[width=9.0cm,height=6.0cm]{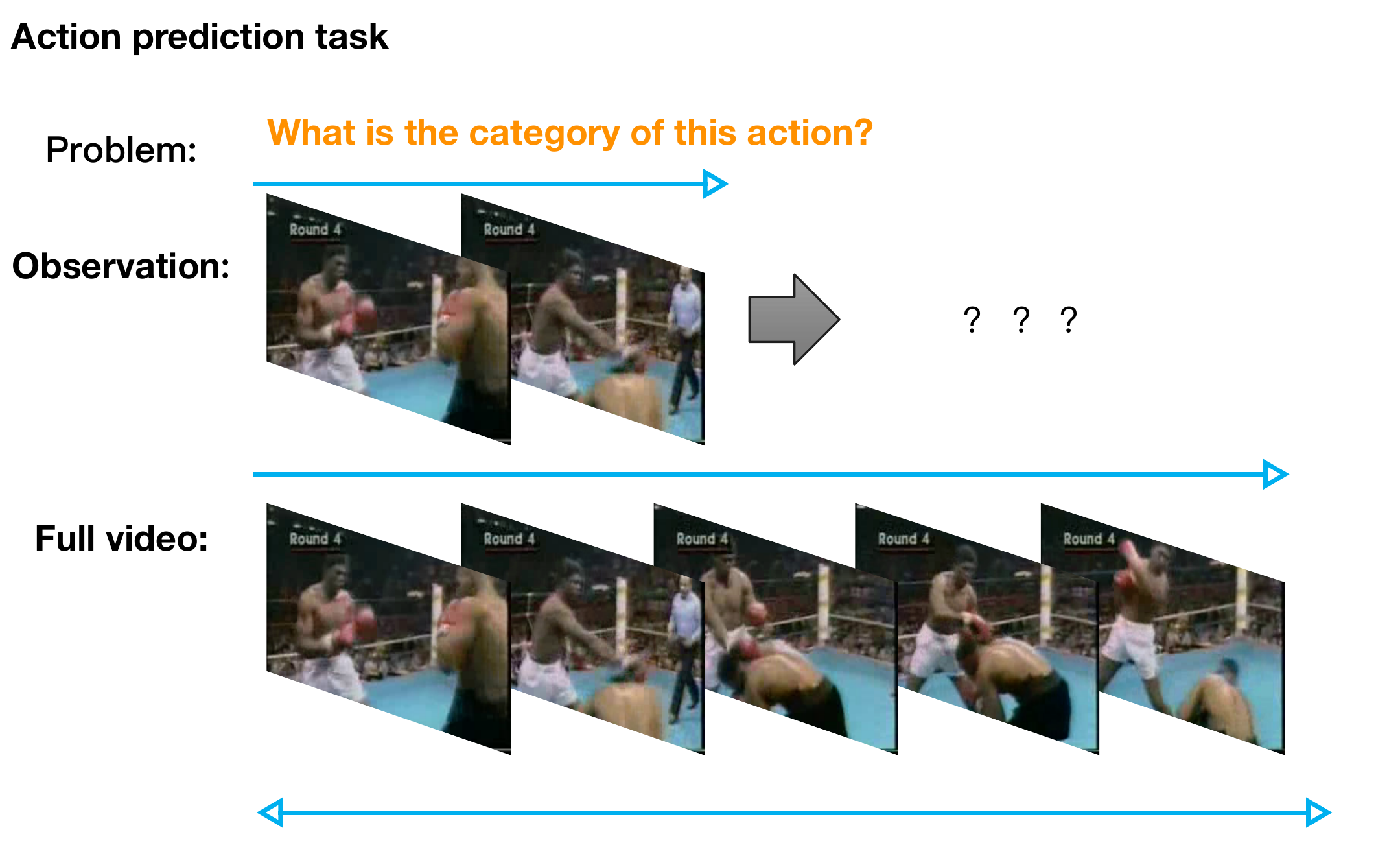}}
\caption{the action prediction problem}
\end{figure}

\par Unlike the action recognition which aims to recognize(or classify) an action under the whole videos, action prediction requires the inference of the action before it occurs, as what in fig 1. Therefore, it is challenging because the computer should recognize the action before the Key Frames happens, which are the climax of an action and of great importance for recognition[1]. Furthermore, if a computer only extracts features from the previous frames, it definitively contains many extra features which are not helpful in predicting what they are doing, for it will extract many features which only exists in that period and have little connection to the future. So, how can we guide the machine to learn something crucial to or have a tight consequence to the future automatically? Inspired by the videos generation and the predictive learning[49], our approach is to make the computer learn certain features which have a strong causal link between the past and the future from the previously known frames. In other words, the occurrence of one incident are under a series of previous incidents and if the computer can recognize those incidents(extract the related feature from those incidents), we would have the high probability to conjecture the upcoming incident correctly. For example, we can inference that the player will throw the basketball by observing what on his hand(basket) and how his arms do(the posture of throwing a basketball) rather than the color and shape of his hair. In this way, we believe that in this scenario the movements of one's arms doing are will be one of the most significant links connecting to the throwing while the color of his hair may not decide the future. The computer is definitely under the same rules. Accordingly, if the machine can extract those incidents' feature which has a great impact on the future, the machine will have the capability to predict the future incidents and even generate the future logically.  
\par The reason why we add generators to our model is that there is no ground truth for those future high influenced features and thus we coerce the machine to generate the future in order to allow it learn more depicting the future.Though depicting the future, we can ensure it will focus on future rather than present.  
\par In this paper, we focus on using Variational AutoEncoder[2](VAE) to learn the future high influenced features from previous frames and exert the those features to infer the future action.Specifically, using two stream 3D-CNN[28](C3D) neural networks to extract feature from previous frames and their temporal information-optical flows.Spatial Pyramid Pooling[3](SPP) is used in both reducing the dimension and extracting muti-scale feature. Then, a generation model, consisted of multi-stream deconvolution neural networks, used to generate the long time future RGB information, short time future RGB information and future temporal information (optical flows) separately. Lastly, after training the generative model, a classifier are used to recognize actions depended on those features.A work flow of our method is shown in Fig. 2.
\par The remainder of this paper is organized as follows. The related works will be illustrated in section II. The detailed model and methodology will be represented in section III. The discussion will be shown in the section IV while the results will be the last section.

\begin{figure*}[!htb]
\centering
{\includegraphics[width=13.8cm,height=4.2cm]{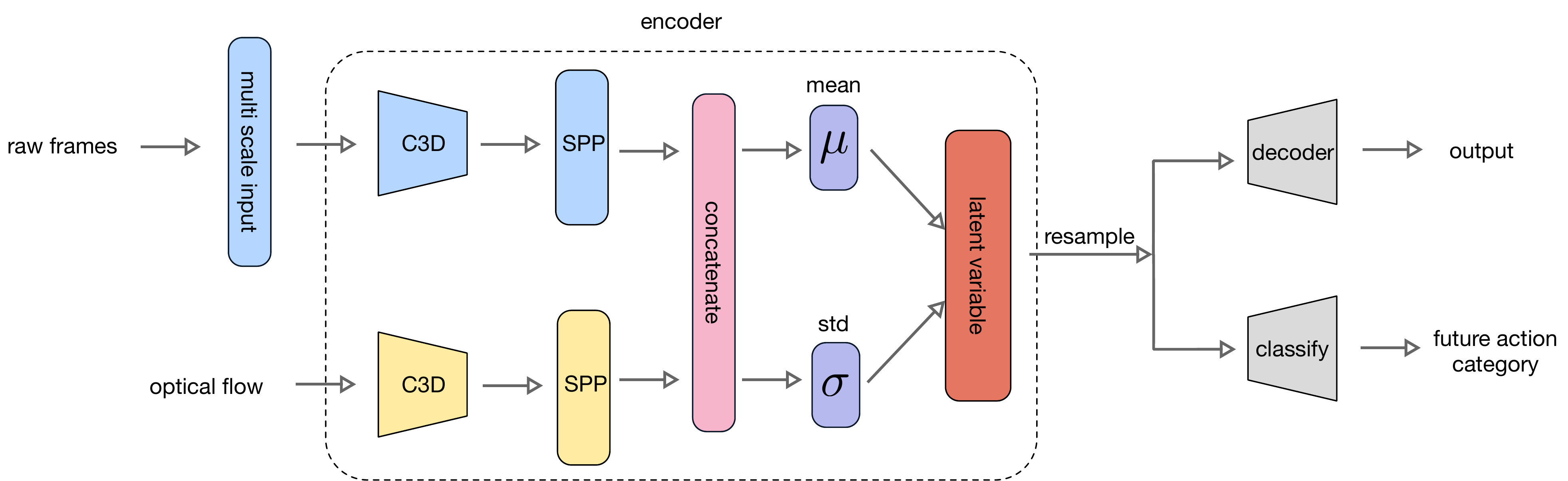}}
\caption{work flow of FL-VAE}
\end{figure*}
\begin{figure*}[!htb]
\centering
{\includegraphics[width=13.8cm,height=5.3cm]{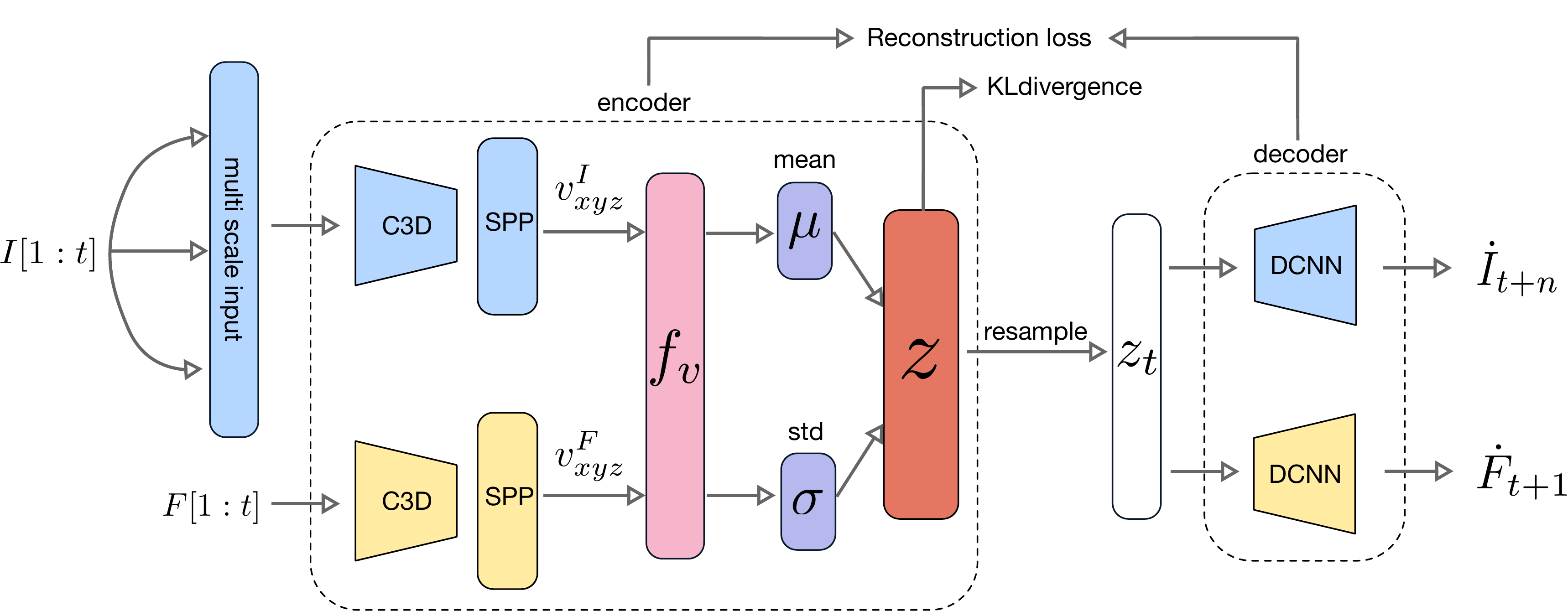}}
\caption{pre-train stage's network architecture}  
\end{figure*}

\section{Related works}		
In this section, we will discuss the existing human Action prediction algorithms and the development of some state-of-the-art structures in our architecture, which including Generative Networks, Two Stream CNN, Spatial pyramid pooling as well as 3D Convolution Neural Network.

\subsection{Existing human Action prediction algorithms}
\par Generally speaking, existing action prediction methods are mainly divided into three parts[4]: the Discriminative model, the Generative model and Deep Learning Network. 
\par The Discriminative model focus on modeling conditional probability distribution, where y are the labels and x are the videos (the same below). It models about target variables conditional on observed variables and less depends on distribution assumption of data. The common used Discriminative model for action prediction are Support Vector Machine[5][6][7][8] and Conditional Random Field[9]. 
\par Generative model pay attention to the modeling full probabilistic model, It wants to learn a full probabilistic model of all variables that generates the data depended on an assumption about the distribution of data. Li et al. used a Probabilistic Suffix Tree (PST) to representing various order Markov Dependencies between action units[10] and Chakraborty et al. treated activity prediction problem as a graph inference problem on Markov Random Field (MRF) where each node is an individual activity[11].
\par Deep Learning methods have been successful in almost all fields and the same as in action prediction and action recognition field.Two main deep learning Network including recurrent neural network(RNN) and CNN.For RNN, Work[12] proposed a novel ranking loss and combines it with classification loss in LSTM network to model activity progression while work[13] combines RNN with Deep Reinforcement Learning. For CNN, work[14] use large-scaled untrimmed videos to train the CNN and predict the future. Works [15][16][17] also introduce CNNs to solve the problems. Moreover, Some scholars[18][19] combine CNN and RNN together to capture both spatial and temporal features. 

\subsection{Two Stream CNN}
\par Since Simonyan et al. proposed the ideas that combining RGB and Optical flow information can increase models? performance[20], more and more action recognition or action prediction methods exert this idea to ameliorate their models. Zhu et al. used traditional local optical flow estimation methods to pre-compute the motion information and RGB information to do the action recognition[31]. Shi et al. add one more stream CNN to capture sequential Deep Trajectory Descriptor in order to add long-term motion information to the model[21]. And Liu et al. Proposed a dynamic gaze CNN which captures both the appearance and motion information from the video and combine it with the original two-stream CNN to improve the correct rate[22].
\par Our works mainly focus on using Two-Stream CNN to capture the present and future spatial , temporal and local spatial features.

\subsection{Generative Networks for Videos Generation}
\par The common Videos generation models include Generative Adversarial Networks(GAN) and VAE. GAN is firstly introduced by Goodfellow et al. in 2014[27], a framework for estimating generative models via an adversarial process. Similarly, VAE, also known as Deep Gaussian Latent Models[2], use a log-likelihood function to reconstruction images.
\par Walker et al. combined both Variational Auto-Encoder(VAE) and GAN to forecast next frames[25]. Vondrick separates the front and back scenes through Foreground-Background Mask in order to increase the models? performance[23]. Liang et al. used dual streams(RGB and optical flow) to predict both next RGB frame and next optical flow frame[24]. Other GAN or VAE networks are also introduced in[29][30]. Our proposed FL-VAE is based on the two-stream C3D and we take advantage of its Variational Inference function to learn the future high connected frames.
 
\subsection{Spatial pyramid pooling and 3D Convolution Neural Network}
\par Spatial pyramid pooling[3] was first introduced by He et al. and widely used in Computer Vision field. Spatial pyramid pooling has two main advantages. Firstly, it reduces the high parameters' dimensions of Convolution Neural Networks. Secondly, since it is the combination of multi-scale max pooling methods, it will extract the features from different scales which is a bit similar to the attention model. Inspired by Zhu et al.[19], we use SPP to extract the features from multi-scale.
\par First introduced by Ji et al[28], 3D Convolution Neural Network, also known as C3D, is one type of Convolution Neural Network which can learn both spatial and temporal features together.It is born with the ability to deal with the video's data. Tran uses large-scale supervised video dataset to do the spatiotemporal feature learning through C3D[32]. Varol et al. use two stream-C3D to combine both the Optical Flow and the RGB information[33].  We also take advantage of C3D spatial-temporal function and use it in our model.

\section{Methodology}
In this section we present our propose FL-VAE architecture. The action prediction task is formula as below: given a partial video $I[1:t], t\in [1,...,T]$, the classifier predict the action category $y$, where $T$ is duration of complete video. In this work, we design a Variational encoder-decoder framework to achieve the action prediction task. The encoder consist of two-stream 3D Convolution neural network with dense layers. And the decoder is deconvolution neural network. As shown in Figure3, in pre-train stage, the encoder encode the input image sequences and fit it to the future high influence action feature $z$, which is denote as latent variable. The resample variables divide into two part($z1$,$z2$) which will be used by two different decoder with distinct generation. Finally the decoder will generate different images (long-term, short-term, past frame or optical flow images). In the train stage, as illustrate in Figure4, we no longer use decoder and instead we use a classificator to classify according to latent variables $z1$,$z2$. Except for that all the processes are identical to the pre-train stage. In the test stage, future information from different variables would concatenate into a action classifier to predict the action category.

\begin{figure*}[!htb]
\centering
{\includegraphics[width=13.8cm,height=5.2cm]{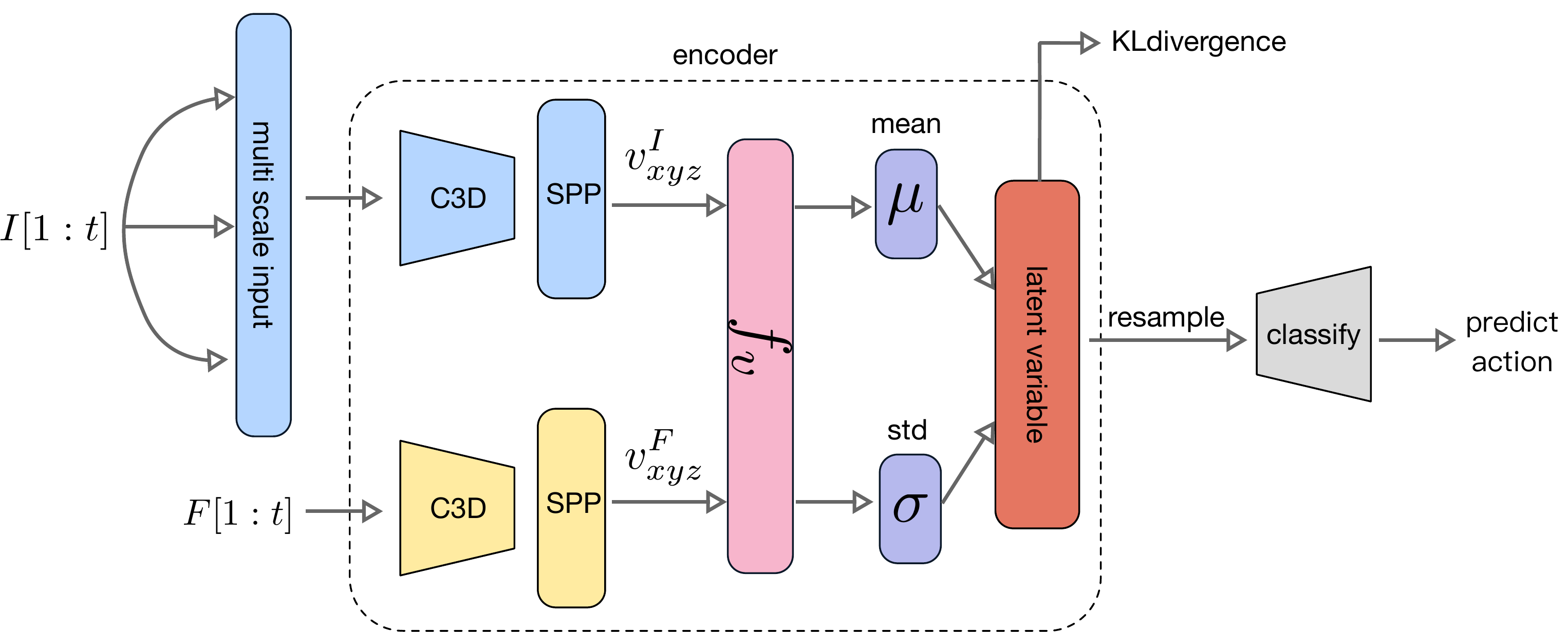}}
\caption{formal train\&test stage's network architecture}  
\end{figure*}

\subsection{Two-stream feature encoder:}
\par Our encoder model consist of a two-stream network: RGB-stream and flow-stream network with three dense layers. The encoder is aim at learning future high influenced features from input sequence. Given the input sequence $S[1:t]$, where S is denote as the input of encoder containing previous optical flow sequence $F[1:t]$ and previous video frames $I[1:t]$, the encoder encode the input sequence $S$ into a stochastic latent variable $z$ and guide the latent variable distribution $z \sim  p(z|S)$ map the truly data probability distribution $q(S)$. $z$ denoted as future action feature.
\par Specifically, in RGB stream, given previous video frames $I[1 : t]$, C3D extract frame-level feature $v_{xyz}^{I}$ from $I$. The output value of C3D network is calculate by:
\[v_{xyz}^{I}=\sigma (b+\sum\limits_{i=0}^{{w}'-1}{\sum\limits_{j=0}^{{h}'-1}{\sum\limits_{k=0}^{{d}'-1}{{{w}_{ijk}}{{k}_{(x+i)(y+j)(z+k)}}}}})\]
Where ${w}'\times {h}'\times {d}'$ is the kernel size of C3D, $\sigma$ is activation function, $b$ is bias, ${{w}_{ijk}}$ is the weight at position $(i, j, k)$ of the kernel and ${{v}_{(x+i)(y+j)(z+k)}}$ is the intensity of the image at position $(x+i, y+j, z+k)$. The mechanism of flow-stream is similar to RGB-stream. Given the previous optical flow sequence $F[1:t]$, the C3D extract frame-level feature and we can get $v_{xyz}^{F}$. And we got fusion frame level feature ${{f}_{v}}$ by concatenating  $v_{xyz}^{F}$ and $v_{xyz}^{I}$. Then ${{f}_{v}}$ would be encode as a mean and variance $\mu ({{f}_{v}})$ and $\sigma ({{f}_{v}})$.  Finally we get the latent variable $z \sim  N(\mu ({{f}_{v}}),\sigma ({{f}_{v}}))$. The goal of encoder is to match $N(\mu (S),\sigma (S))$ as closely as possible. In most circumstances, the distribution of image data satisfies Gaussian distribution, so $N(\mu (S),\sigma (S))$ can be regarded as $N(0,G)$. Where G is a N$\times$N identity matrix. During the training stage, latent variable $z$ keep trying to fit $N(0,G)$.
\par In addition, to improve the performance of encoder, we exploit different strategy to construct the convolutional descriptors. Spatial Pyramid Pooling(SPP) layer could help CNN extractor extract scale-invariant local descriptor. We employ SPP on the last layer of C3D both in RGB and optical flow stream. To extract as much as useful information from input video frames I, we apply random crop size method. The width and height of frame would randomly select from $\{$$R1,R2,R3,R4$$\}$(${{R}_{i}}$ is positive integer) and then stack with the original video frame to feed in our model. This method could create more training data and make full use of the input RGB information.
\subsection{Multi kinds of decoder:}
\par As mentioned above, we apply decoder model in pre-train stage. The effect of decoder is reconstruct the future information while given the input $z$. The decoder network has two roles. First, we couldn't evaluate the quality of future high influenced features $z$ directly. In the contrast, we evaluated quality indirectly by future spatial-temporal information which  generated by decoder. Second, the existence of decoder could boost the performance of representation learning  because it could aid encoder capture more information about future action. 
\par To exploit the impact of different kinds future information, we develop two kinds decoder model respectively: future optical flow generator and future video frame generator.\\
\textbf{a. future video frame generator}
\begin{figure}[!htb]
\centering
{\includegraphics[width=9.16cm,height=4.17cm]{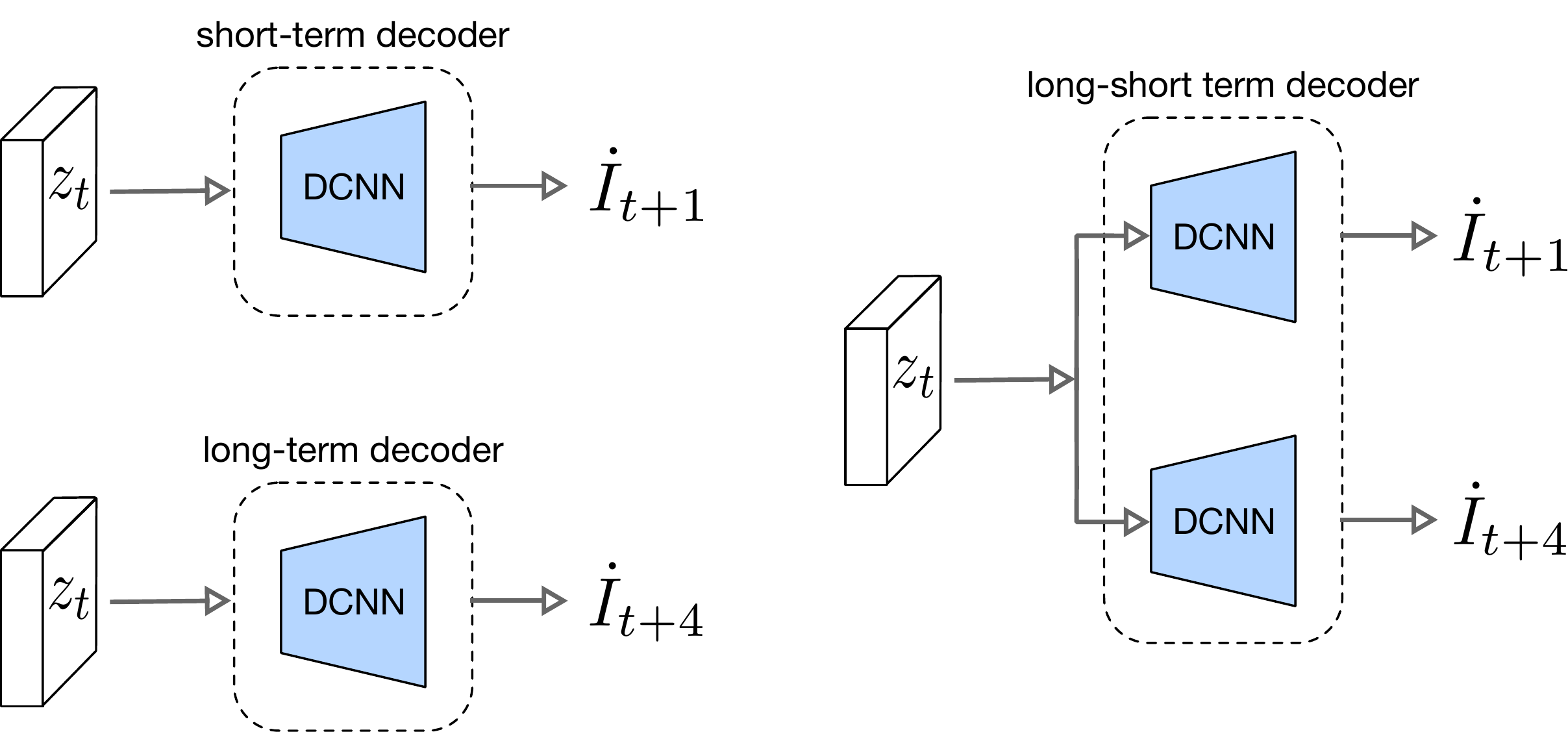}}
\caption{different kinds of future video frame decoder}  
\end{figure}

\par As shown in Figure 5, we model two kinds of future video frame generator: short-term generator and long-term generator. Given the resampled  latent variable $z_t$, the future video frame generator can generate future video frame ${{\dot{I}}_{i+n}}$. For short-term generator, $n$ is set as 1 however $n$ is set as 4 in long-term generator. Then we assess the quality of latent variable by Mean-square prediction/reconstruction error:
\[{L_{R\_I}}={{{({{I}_{t+n}}-{{{\dot{I}}}_{t+n}})}^{2}}}\]
Where ${{I}_{t+n}}$ is ground true future sequence. ${{L}_{R}}$ is denote as short-term future error when n=1 and it is regard as long-term future error for n=4. 
\par Further, we build a long-short term future video generator with the combination of short-term generator and long-term generator.  And the error function is modified to:
\[{{{L}'}_{R\_I}}={{{({{I}_{t+1}}-{{{\dot{I}}}_{t+1}})}^{2}}}+{{{({{I}_{t+4}}-{{{\dot{I}}}_{t+4}})}^{2}}}\]
Where ${{{L}'}_{R\_I}}$ is denote as long-short term future error function. To minimize the ${{{L}'}_{R\_I}}$, encoder need to capture more high connect future feature which is crucial for action prediction. The settings of long-short term future video generator could significantly improve the performance of representation learning.\\
\textbf{b. future optical flow generator}
\par The network structure of future optical flow generator are similar to the future video frame generator. However, we only model the short-term future optical flow generator. So the future optical flow generator only generate future optical flow ${{\dot{F}}_{i+1}}$ when given the resampled latent variable $z_t$. Same as future video frame generator, we evaluate the quality of latent variable by Mean-square prediction/reconstruction error:
\[{{L}_{R\_F}}={{{({{F}_{t+1}}-{{{\dot{F}}}_{t+1}})}^{2}}}\]

\subsection{Model optimization:}
\par To achieve better performance, we train our model with multi loss function. We utilize three object function shown as follow: \\
\textbf{a}.  A pixelwise reconstruct error function, which is calculate as:
\[{{L}_{R}}={{{({{S}_{t+n}}-{{{\dot{S}}}_{t+n}})}^{2}}}\]
  where $\dot{S}$ is generated future sequence containing ${{\dot{I}}_{i+n}}$ or ${{\dot{F}}_{i+n}}$through decoder.\\
\textbf{b}.  The object function of VAE is define as
\[{{L}_{VAE}}=KL(p(z|S)||q(S))\]
where $KL$ is the Kullback-Leibler divergence that penalizes deviation of the distribution of the z from the $N(0,G)$.\\
\textbf{c}. To avoid over-fitting, we employ L2 regularize method. Let ${{\mu }_{T}}$ denote the collection of model network's weight variable, we can calculate the L2 regularize loss as below:
\[{{L}_{l2}}=\frac{1}{2}||{{\mu }_{T}}||\]
The final object function $L$ is obtain by fusing these three loss:
\[L={{\lambda }_{1}}{{L}_{R}}+{{\lambda }_{2}}{{L}_{VAE}}+{{\lambda }_{3}}{{L}_{l2}}\]
Where ${{\lambda }_{i}}$ is scale factor. During the model training process, the model parameter would adjust and update by minimizing $L$.

\subsection{Action predictor:}
\par After encoder training completed, the latent variable $z \sim p(z|S)$ would fit the truly future action probability distribution. Then we can recognize future actions by resampling latent variable $z$. For action predictin task, we implement a classifier through a dense network. The input of classifier is resampled $z_t$. The output layer of classifier is N-way softmax layer to predict the probability distribution over N different actions:
\[{{y}_{i}}=\frac{\exp ({{{{y}'}}_{i}})}{\sum\nolimits_{k=1}^{N}{\exp ({{{{y}'}}_{k}})}}\]
where ${{{y}'}_{i}}=\sigma (w*{{z}_{t}}+b)$. The classifier is trained by minimizing $L_{cla}$:
\[{{L}_{cla}}=softmax\_loss({{y}_{i}},{{l}_{i}})\]
Where ${{l}_{i}}$ is the ground true label.

\subsection{Network setup:}
\par The details of the FL-VAE encoder network are shown in Figure4. The C3D contain 4 three-dimension convolution layers. We apply dropout layer between each convolution layer and utilize L2 regularize method to avoid over-fitting. Each convolution layers' activate function is Relu. The bins of SPP net is set as [4,2,1]. We use two dense layers to encode $\mu $ and $\sigma $, and a dense layer to sample $z$. All the dense layers have 12 output units.
\begin{figure*}[!htb]
\centering
{\includegraphics[width=14.0cm,height=6.0cm]{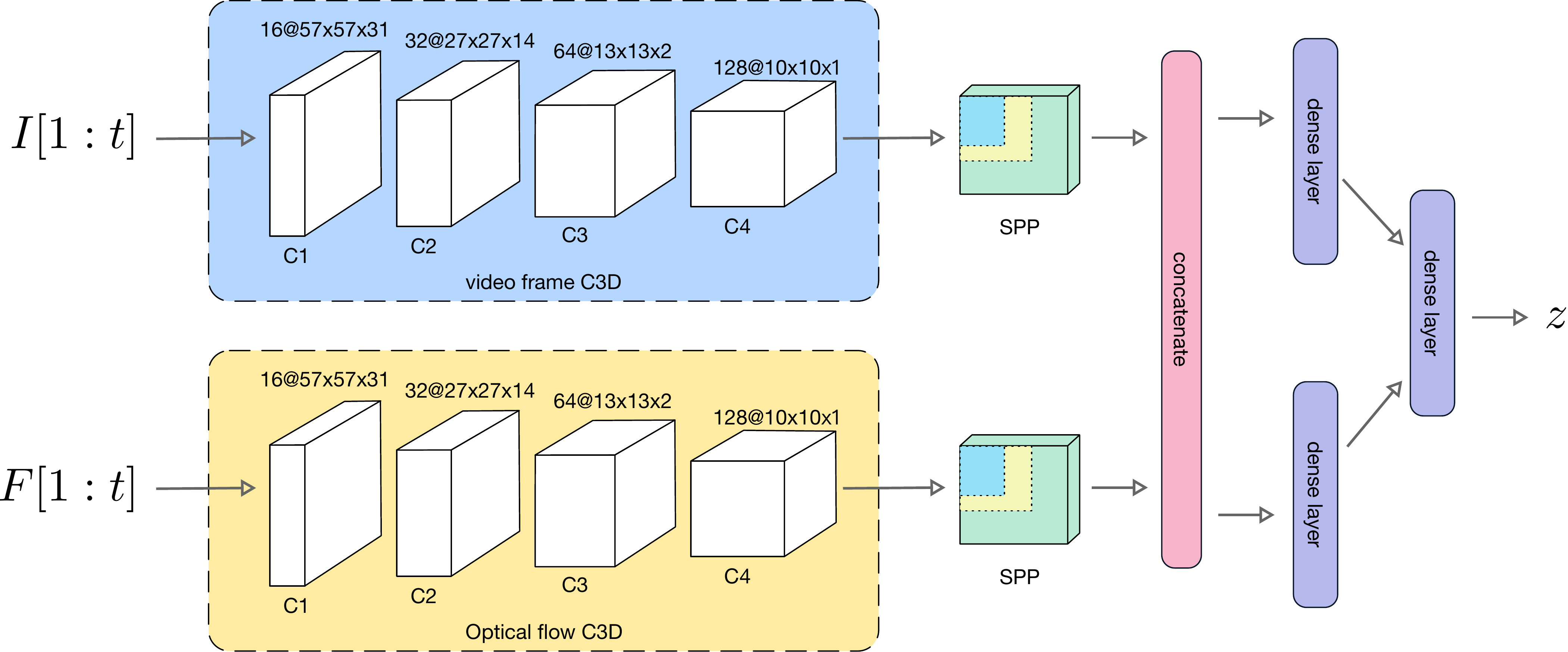}}
\caption{the detail of encoder model}
\end{figure*}	
\par The detail of decoder network is shown in Figure5. All the decoder contain 5 deconvolution layers. In each layer we utilize Relu activate function, dropout layer and L2 regularize method. The action predictor network is implement as one dense layer containing 10 output units.  
\begin{figure*}[!htb]
\centering
{\includegraphics[width=14.0cm,height=3.5cm]{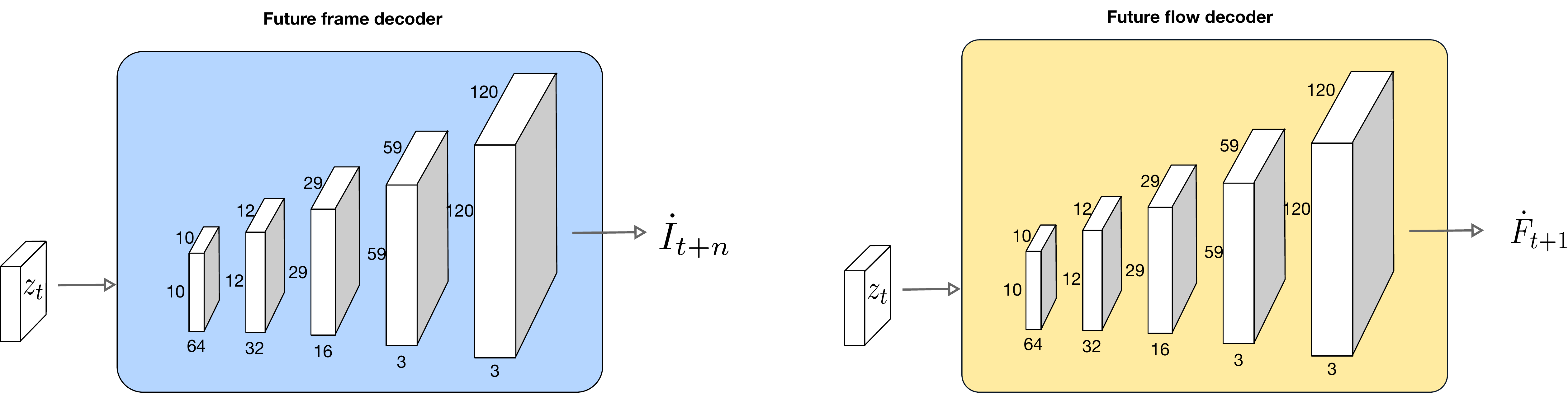}}
\caption{the detail of decoder model}
\end{figure*}

\section{Experiment}
\subsection{Datasets}
\par We implement pre-train process on the UCF101 datasets[44]. UCF101 is a benchmark for action recognition which contain 13k clips video. The total number of actions category is 101. We use this as unsupervised learning dataset in pre-train stage.
\par We experiment on the UT-Interaction dataset (UTI) [34]. The UTI contains 6 classes of human-human interactions videos: shake-hands, point, hug, push, kick and punch. Ground truth labels for these interactions are provide. Each class contain 10 video. We use UTI$\#$1 as the train and test dataset.
\subsection{Implement detail}
\par The implement of purpose model is based on Tensorflow[35]. We utilize an Adadelta optimize algorithm[36] to update the network variable. The original learning rate is initialized as 1 and the decay rate is set as 0.95. All the weights are initialized by truncated normal initializer with mean 0 and standard deviation 0.1 while bias are initialized with mean 0 and standard deviation 0.01. All the nets are with L2 regularizer and their weight-decays are set to 0.001.
\par In this work, we use optical flow to represent the temporal feature of a consequence of a videos. There are several algorithms to calculate the optical flow[34][35][36]. Since our goal is to extract the partial temporal information. So, the dense optical flow is our best choice. After comparing, we choose TV-L 1 optical flow[37] as our model?s optical flow.
\par Two Data Augment method used in model are Random Clipping and Multi-scale Cropping. For multi-scale cropping, original video frames would randomly crop their size choose from {320,360,400}, and then stack with original video sequence to input to encoder. All the input training frames would resize to 120x120 before feed to the network. For random clipping method, each video is clipped into a series of frames and we choose the frame randomly but in same order. In our experiment, we choose frame length is 64. The optical flow is compute by TVL1 algorithm implement in OpenCV package.
For all the experiment, we adopt leave-one cross validation method to measure the performance of the datasets, like the experiment mentioned in[37]. That is, for each set, 3 sequences of segmented videos (45) are used for training and the remaining 15 videos are used for testing.
\par The complete process of training and testing are achieve on Titan X GPU with Xeon CPU.
\subsection{Performance on UT-I datasets}
\textbf{Part1: compare of different training parameters}
\par We first examine the effect of different training trick. In this part we want to find which parameters could improve the performance of video encoder. We investigate the advantage of pre-train method by pre-training model on UCF101 datasets and then fine-tuning on UTI datasets. To find out the impact of data augment, we utilize random clipping and multi scale method. To decrease the impact of over-fitting, we apply two different dropout probability, one is 0.9 and the other is 0.5. Finally, we evaluate the performance by the predicting accuracy. The result are summarized in table 1.
\begin{table*}[htbp]
  \centering
    \begin{tabular}{|c|c|c|c|c|c|}
    \toprule
    Pre-training in UCF101 & Random clipping & Crop size & Dropout probability &  Multi scale method & Accuracy \\\midrule
    Yes   & Yes   & 64    & 0.5   & No    & 56.7\% \\\midrule
    No  & Yes   & 64    & 0.5   & No    & 20.0\% \\\midrule
    Yes   & No    & 64    & 0.5   & No    & 53.3\% \\\midrule
    Yes   & Yes   & 120   & 0.5   & No    & 58.3\% \\\midrule
    Yes   & Yes   & 64    & 0.9   & No    & 56.7\% \\\midrule
    Yes   & Yes   & 120   & 0.9   & No    & 58.3\% \\\midrule
    \textbf{Yes}   & \textbf{Yes}   & \textbf{120}   & \textbf{0.9}   & \textbf{Yes}   & \textbf{63.3}\% \\\bottomrule
    \end{tabular}%
    \caption{performance comparisons of different method. We compare different tricks in our experiments}
  \label{tab:addlabel}%
\end{table*}%

\par As shown in the Table 1, we get the best performance with pre-train method, random clipping policy and multi-scale method, 120 crop size. The usage of pre-train method can significantly improve the model performance.It?s mainly because traditional classifier should find a way to coverage but itself while pre-train method play a role as a guider to lead the classifier to a road more easily to coverage and thus reduce the time. Under the same condition, we gain nearly 40$\%$ with the pre-train method and the model have far better performance than training from scratch. In addition, we gain above 4$\%$ with random clipping meaning that it is a useful data augment trick. However, we get same predict accuracy with different dropout probability, means that the choice of high dropout probability have less impact in our experiment than others. Although multi scale method the accuracy reach a high mark, the high-computation and time consuming are unbearable to some extend. Therefore, we do not use this method commonly in next experiment. Since pre-train method, low dropout, random clipping and 120 crop size method reveals one of the best results in the experiment, We utilize them for all remaining experiments in this paper. 
\\
\newline
\textbf{Part2: The performance of different encoder and decoder}
\par To exploit the optical flow effect on encoder model training, we augment an optical flow encoder and compare the result with single RGB input. The impact of optical flow in future representation learning is shown in Table2. As illustrate in the table, with optical flow input, the model performance achieve a incredible improvement. With the usage of optical flow, the improvement of accuracy could achieve more than 20$\%$. This suggests that the temporal information which contain useful motion feature could effectively improve the model learning performance. We can conclude that the optical flow play an important role in learning future representation for action prediction. 

\begin{table*}[htbp]
  \centering
    \begin{tabular}{|c|c|c|c|}
    \toprule
    \multirow{2}[2]{*}{Encoder} & \multicolumn{1}{c|}{\multirow{2}[2]{*}{Decoder}} & \multicolumn{1}{c|}{Accuracy } & \multicolumn{1}{p{5em}|}{Accuracy } \\
    \multicolumn{1}{|c|}{} &       &    \multicolumn{1}{c|}{Ratio 0.5}    & \multicolumn{1}{c|}{Ratio 1} \\
    \midrule
    \multirow{3}[8]{*}{RGB} & \multicolumn{1}{c|}{Short term RGB} & 63.3\% & 68.3\% \\
\cmidrule{2-4}    \multicolumn{1}{|c|}{} & \multicolumn{1}{c|}{Short term RGB + long term RGB} & 66.7\% & 71.7\% \\
\cmidrule{2-4}    \multicolumn{1}{|c|}{} & \multicolumn{1}{c|}{Short term RGB + future optical flow} & 56.7\% & 61.7\% \\
    \midrule
    \multirow{4}[10]{*}{RGB + flow} & \multicolumn{1}{c|}{Short term RGB + future optical flow} & 78.3\% & 81.7\% \\
\cmidrule{2-4}    \multicolumn{1}{|c|}{} & \multicolumn{1}{c|}{Short term RGB} & 83.3\% & 86.6\% \\
\cmidrule{2-4}    \multicolumn{1}{|c|}{} & \multicolumn{1}{c|}{Short term RGB + long term RGB} & 86.7\% & 88.3\% \\
\cmidrule{2-4}    \multicolumn{1}{|c|}{} & \multicolumn{1}{c|}{Past RGB} & 83.3\% & 90.0\% \\
    \midrule
    \textbf{RGB + flow} & \multicolumn{1}{c|}{\multirow{2}[2]{*}{\textbf{Short term RGB + long term RGB}}} & \multirow{2}[2]{*}{\textbf{93.3\%}} & \multirow{2}[2]{*}{\textbf{95.0\%}} \\
    (\textbf{with multi scale}) &       &       &  \\
    \bottomrule
    \end{tabular}%
   \caption{the performance comparison of different encoder and decoder.}
  \label{tab:addlabel}%
\end{table*}%

\par To evaluate the effect of the generated future hand-crafted feature, we compare our model with several kinds of the decoder. There is two kinds decoder component in our network setting: future RGB decoder and future optical flow decoder. Specifically, RGB decoder contains a normal future RGB generator(short-term RGB) and a long-term RGB generator. The comparison result is shown in table 2. There are four conclusions:
\par Firstly, a single generate future hand-craft feature could help boost the predicted accuracy. With the usage of RGB decoder, the accuracy could boost at least 3.3$\%$. This is mainly because the existent of RGB decoder could help encoder capture more input video feature with a high causality to the future. 
\par Secondly, the fusion decoder of future RGB and optical flow decoder could not boost the performance. On the contrary, this method gets worse performance than single future RGB decoder.This occurs may owe to the negative effect of overabundance features and we will discuss it further in the next section.
\par Thirdly, with long-short term future RGB generator, we can achieve the best improvement. This is mainly because that the long-short term future RGB generator could force the encoder to learn more future high connected feature from input video.
\par Last but not least, all the model performs better in ratio 1 than in ratio 0.5. Nevertheless,  the longer frame it generates, the less increment it has.E.g. the RGB+Flow encoder with long-short term RGB decoder rises less than Short term RGB decoder from ratio 0.5 to 1. This mainly because when the keyframe reaches, the most important feature is known. The action prediction method degenerates into the action recognition method. So, the future generation network loses its effect. We will discuss it later.
In particular, we can obtain the further performance with the multiscale method. This demonstrates the effectiveness of propose FL-VAE.
\\
\newline
\textbf{Part3: compare of different observe ratio}
\begin{figure}
{\includegraphics[width=9.0cm,height=6.0cm]{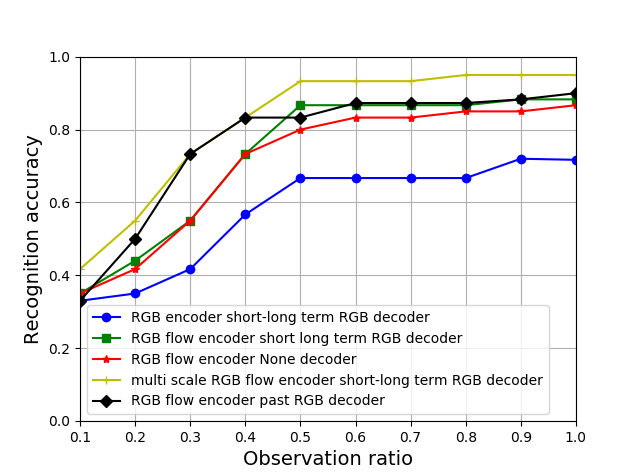}}
\caption{Prediction results on UTI $\#$ 1}
\end{figure}
\par Fig 6. reveals the details of action prediction when tested on UTI$\#$1. We ferret out that at first they are almost the same. That is natural because a model without any train can also get an accuracy about 20$\%$ and thus the low accuracy can not tell how the quality of the model. But when the ratio increases, their performances become different. The model with the multi-scale method raise the fastest from ratio 0.2 to 0.5. That shows the effect of data augment method. Something interesting is that after 0.5 ratio all the models do not raise their performances dramatically. That is because the fastest ascending regions are key frames region. And after this region, less useful information will exists in the videos and naturally their performance raise less. The two-stream encoder with long-short term RGB decoder at first overcome the two-stream encoder with past RGB decoder but finally loses is mainly because this kind of future prediction models are not so suitable when there is nothing can be predicted(full observation). 
\\
\newline
\textbf{Part4: comparing with other state-of-the-arts}
\begin{table}[htbp]
  \centering
  
    \begin{tabular}{|l|c|c|}
    \toprule
    Method & \multicolumn{1}{p{5em}|}{Half observation(\%)} & \multicolumn{1}{p{5em}|}{Full observation (\%)} \\
    \midrule
    Dynamic BoW[31] & 70    & 85 \\
    \midrule
    Integral BoW[31] & 65    & 81.7 \\
    \midrule
    Cuboid+SVMs[37] & 31.7  & 85 \\
    \midrule
    Kong [7] & 78.33 & 95 \\
    \midrule
    Hierarchical movemes[45] & 83.1  & 88.4 \\
    \midrule
    Ke[15] & 83.33 & - \\
    \midrule
    Xu[46] & 70    & 80 \\
    \midrule
    Poselet[6] & 73.3  & 93.3 \\
    \midrule
    AAC[38] & 91.67 & 96.67 \\
    \midrule
    \textbf{Our method} & \textbf{93.3}  & \textbf{95} \\
    \bottomrule
    \end{tabular}%
    \caption{Activity prediction performance on UTI $\#$ 1 dataset}
  \label{tab:addlabel}%
\end{table}%
\par Our model compare with the stat-of-the-art models:
including 1)Bag-of-words based methods: Dynamic BoW and Integral BoW[31].2)SVM method: Cuboid+SVMs[37] 3)Max margin action prediction method: Kong[7] 4)hierarchical representation method:Hierarchical movemes[45] 5)deep temporal features method:Ke 6)learning combinatorial sparse representations:Xu[46] 7)Poselet key-framing model: Poselet[6] and 8)discriminative patch based method: AAC[38].
\par From table 3 we can know that, our best method achieves favorable performance compared to other methods. The accuracy of Half observation is 93.3$\%$, higher than the previous stat-of-the-art model:AAC whose accuracy is 91.6$\%$. Although the full observation accuracy of our model is not beat the best method- also AAC and its accuracy is 96.67$\%$ at this time, our model ,with 95$\%$ accuracy, is near to it. The reason why our method does not overcome AAC is mainly because the model is suitable for prediction rather than action recognition. Since it learns the future high connected features, our method will concentrate more on prediction.
In all, our results are higher than most of the traditional methods as well as latest deep learning methods.
\\
\newline
\textbf{Runtime analysis}
\begin{table*}[htbp]
  \centering

    \begin{tabular}{|l|r|}
    \toprule
    Model  & \multicolumn{1}{c|}{Runtime} \\
    \midrule
    RGB encoder & \multicolumn{1}{r|}{\multirow{2}[2]{*}{0.6 day pre-train + 0.2 day train}} \\
    Short term RGB decoder &  \\
    \midrule
    RGB + flow  encoder & \multicolumn{1}{r|}{\multirow{2}[2]{*}{0.6 day pre-train + 0.2 day train}} \\
    Short term RGB decoder &  \\
    \midrule
    RGB + flow  encoder & \multicolumn{1}{c|}{\multirow{2}[2]{*}{0.4 day train}} \\
    None decoder &  \\
    \midrule
    RGB + flow encoder with multi scale & \multicolumn{1}{c|}{\multirow{2}[2]{*}{About 3 days train}} \\
    Short term RGB + long term RGB decoder &  \\
    \bottomrule
    \end{tabular}%
      \caption{Runtime of each typical model in UT\#1}
  \label{tab:addlabel}%
\end{table*}%
\par  Pre-training on UCF101 takes 0.6 day for 3000 epoch. The usage of pre-train method reduce the training time from 0.4 to 0.2 day. Though pretrain itself wastes many time, it does speed up the convergence and low down the probability of over-fitting. As we mentioned above, using the multi scale method, the consume time would increase exponentially. The total consuming time is from 0.8 day to 3day. But it accelerates the convergence process and can reach the best result. Our propose method is more effective.

\subsection{Performance on UCF101 datasets}
\textbf{Part1: The performance of different encoder and decoder}
Our model performance result on UCF101 is shown in table 4. Same as UTI's result, with RGB-flow encoder model we can obtain the better performance than single RGB encoder architecture. The accuracy improve nearly 20$\%$ with optical flow input.  We also evaluate the model performance without pre-train stage which denote 'None decoder' in table 4. It demonstrate that accuracy would reduce by nearly 10$\%$ without decoder model, which prove the effectiveness of decoder. However, unlike the UTI’s result, in UCF101 experiment long-short term decoder has similarity accuracy with short-term decoder. In addition, in UCF101 datasets the accuracy od Past-RGB decoder is close to short-term decoder and long-short term decoder. This is mainly because that many of video sequence in UCF101 are low prediction(could be predicted with less than 50$\%$ observation), so the effect of long-term decoder is not significant. Plus, we do not use multi-scale method because using this method in UCF101 dataset is too time-consuming.
\begin{table}[htbp]
  \centering
    \begin{tabular}{|c|c|c|c|}
    \toprule
    \multicolumn{1}{|c|}{\multirow{2}[2]{*}{Encoder}} & \multicolumn{1}{c|}{\multirow{2}[2]{*}{Decoder}} & \multicolumn{1}{c|}{Accuracy } &\multicolumn{1}{c|}{Accuracy}  \\
          & \multicolumn{1}{c|}{} &    Ratio 0.5   &  \multicolumn{1}{c|}{Ratio 1} \\
    \midrule
    \multicolumn{1}{|c|}{\multirow{2}[2]{*}{RGB}} & \multicolumn{1}{c|}{Short term RGB + } & \multicolumn{1}{c|}{\multirow{2}[2]{*}{62.0\%}} & \multicolumn{1}{c|}{\multirow{2}[2]{*}{62.0\% }} \\
     	& \multicolumn{1}{c|}{long term RGB} & & \\
    \midrule
          & \multicolumn{1}{c|}{None}  & 71.2\% & \multicolumn{1}{c|}{75.0\%} \\
\cmidrule{2-4}          & \multicolumn{1}{c|}{Short term RGB + } & \multirow{2}[2]{*}{83.3\%} & \multicolumn{1}{c|}{\multirow{2}[2]{*}{86.6\%}} \\
    		 	 &        \multicolumn{1}{c|}{long term RGB} & & \\
\cmidrule{2-4}    \multicolumn{1}{|c|}{\multirow{1}{*}{RGB + flow}} & \multicolumn{1}{c|}{Short term RGB} & 86.0\% & \multicolumn{1}{c|}{86.8\%} \\
\cmidrule{2-4}          & \multicolumn{1}{c|}{Past RGB} & 85.5\% & {87.9\%}  \\
    \bottomrule
    \end{tabular}%
      \caption{model performance on UCF101 datasets}
  \label{tab:addlabel}%
\end{table}
\\
\newline
\textbf{Part2: numerical result of Mean-square error}
Table 5 show the result of the average single normalization image's Mean-square error(MSE) in pre-train stage. The small numerical value demonstrate that the generated future video frame and future flow is close to ground true, which means that the latent variable $z$ fits future high inuenced features well. Short-term RGB decoder’s MSE is closed to that of Past RGB decoder, which means that reconstruction and short-term generation have the error. But short-term generation outperforms reconstruction in prediction. This result strength the hypothesize that under the condition of the same network, generation does learn certain features which contain more useful information to predict the future. 
\par Moreover, It is no surprise that long-term decoder’s MSE is large, for the the more deeper in the future, the less information we can grasp. 
\begin{table}[htbp]
  \centering  
    \begin{tabular}{|c|c|}
    \toprule
    Decoder & MSE \\
    \midrule
    Short-term RGB decoder & 0.07426 \\
    \midrule
    Long-term RGB decoder& 0.08143 \\
    \midrule
    Past RGB decoder & 0.07324 \\
    \midrule
    Optical flow decoder & 0.1403 \\
    \bottomrule
    \end{tabular}%
    \caption{MSE result in UCF101 datasets}
  \label{tab:addlabel}%
\end{table}
\\
\newline
\textbf{Part3: comparing with other state-of-the-arts}
\par Also, we compare our method with stat-of-the-art model in UCF101, including 1): C3D+SVM and C3D+X2SVM[42] 2): IBOW[31], 3): MTSSVM[47] and 4):Deep Sequential Context Networks[41]. From table 6 we can conclude that our method get better performance compared to the other result. Almost compare method except Kong[41] achieve nearly 80$\%$ prediction accuracy with full observation. In comparison, our method achieve 86$\%$ prediction accuracy with only half observation due to the effect of future high inﬂuenced features learned by Predictive learning model. The half observation accuracy of our method is higher than Kong[41] whereas lower than Kong[41] with full observation. This is mainly because that our Predictive learning model will concentrate more on prediction rather than recognition. In all, our results are higher than most of the traditional methods as well as latest deep learning methods.

\begin{table}[htbp]
  \centering
    \begin{tabular}{|c|c|c|}
    \toprule
    Method & \multicolumn{1}{p{5em}|}{Half observation(\%)} & \multicolumn{1}{p{5em}|}{Full observation (\%)} \\
    \midrule
    C3D+SVM[41] & 80.3\% & 81.6\% \\
    \midrule
    C3D+X2SVM[41] & 81.0\% & 83.0\% \\
    \midrule
    IBOW[31] & 74.3\% & 75.8\% \\
    \midrule
    MTSSVM[47] & 82.4\% & 82.8\% \\
    \midrule
    MSSC[48] & 61.3\% & 61.8\% \\
    \midrule
    Kong[41] & 85.8\% & 87.6\% \\
    \midrule
    Our method	& 86.0\%	&86.8\%\\
    \bottomrule
    \end{tabular}%
      \caption{activity prediction performance on UCF101 datasets}
  \label{tab:addlabel}%
\end{table}

\section{Discussion}
\par As what we repeatedly emphasized above, the aim of our method is to use unsupervised learning aids action prediction. And two kind of unsupervised learning we use: One is reconstruction, the other is Generation. 
\par The reconstruction method has been used by Qiu et al., who takes advantage of VAE as a replacement to Fisher Vector(FV)[40]. This FV-VAE method is similar to past RGB decoder our model, which learns distinct two videos automatically while the traditional FV uses Gaussian Mixture Model (GMM) to measure the similarity between the two sets. Similarly, Bütepage et al. adopt the same idea- using deep representation learning to help aid human motion classification[39].
\par For Generation aids human action prediction, as far as we know, there is no any research uses this method to help the action prediction directly. But some scholars do this experiment in passing. Liang et al. Use this ideas but just mainly focus on the videos generation through Dual Motion GAN[24] without prediction. 
\par Comparing with the state-of-the-art methods, we find that in fact many methods owns the same ideas. Extracting the features and using them for aids classification or cluster. Some use the statistical machine learning methods, like BOW and SVM[37]. Others choose to learn the features and classification automatically, like AAC[38] and DSCN[41]. Our method is the same as learn feature extraction aromatically but we focus on future high connected features.  
\par The followings we will focus on some specific details:  \\
\newline
\textbf{A. Why using optical flow as decoder does not boost the accuracy well?}
\par From the experiment mentioned above, we find that adding hand-crafted features(like the optical flow) to the decoder do not boost the accuracy of the model well while to the encoder increase the models performance dramatically. So, why this happens? 
\par In our point of views, since the Back Propagation from one feature will influence the others, we believe that the complicated features to reconstruct may harass the learning of the latent variables. When forcing the Variant AutoEncoder to learn the feature that is our human-made as well as the originally Image together, there may be too complicated for the Variant AutoEncoder. So, if we consider in this way, it will not surprise us that decoding more descriptors will make the predict performance worse.\\
\newline
\textbf{B. Which generator is better? the past or future}
\par As we emphasized before, After doing the experiments above, we find that generative model does improve the performance of the model. But there is no significant difference in the generative model between past generative model and the future generative model. There is one possible reason to explain it. 
\par In UT dataset, it is the position of the key frame that is less than half of the videos.e.g, if the whole video is 200 frames long and the key frame is with the onset of the 90th frame. So, the significant features contained in videos are exposed to the past generative model and thus it can improve the accuracy well.We examine the videos and find out that the result does conform to our hypothesize. 
\par Plus, the UCF-101 also Strengthen this hypothesize. Since In UCF-101, more than 70$\%$(71 out of 101) videos are instantly predictable (the video can be predicted after only observing the beginning 10$\%$) or early predictable (the video can be predicted if the beginning 50$\%$ portion of the video is observed) according to a research[41]. Therefore, since most of the videos get into key-frames after half observation, the advantage of learning the high connection future feature might not be so useful in this dataset.
\par Also, we are willing to discuss more the future generative model-why do this model improves its accuracy more slowly when the ratio gets nearer to 100$\%$. It is also explained by the location of the Key Frames. Since the Key Frames already ended, the future high influenced features are no longer useful. So naturally, the past generator get the same results as the future generator. Moreover, we can find that the future generator not only learns the future high influenced features but also learn the features for present action recognition, for the future generative model do well in prediction its performance when the ratio is 100$\%$. \\
\newline
\textbf{C. Which tricks is cost-effective: Data Augment, high drop-out or Two Stream Network?}
\par Data augment is one of the most effective methods to improve the performance. Despite increase the training time, Random clipping and multi-scale cropping create more data and thus diminish the over-fitting of the model.
\par High dropout does not reveal a strong improvement of performance as mentioned in [33]. Perhaps it is because its effect is mainly based on the choice of the dataset or it is the pre-train method that replaces.
\par Combining the Optical Flow and the RGB information together through two-stream C3D is a super effective method in boosting the accuracy. In fact, the success of the model is principally contributed to two things: one is the future generative network while the other is the two-stream network.
\par In all, the contribution to the performance can be ranked as follows: future generative network, two-stream C3D, Data augment, high dropout.\\

\section{Conclusion}
\par We have represented the Future Representation Learning Variational AutoEncoder(FL-VAE).The main goal of FL-VAE is to learn to the certain future-important features, which have  tight connections with future scenarios, automatically according to past known frames and uses these features to do the action prediction.Then, a variety of experiments were done to verify our claim and ferret out that the model with RGB and optical flow as input and long future RGB and short future RGB of output reach the highest mark among the models we proposed. Moreover, comparing with the state-of-the-art algorithms, our model performs well and thus prove that learning the future-important features can help increase the performance of the model.  
\par Our future work is as follows. Firstly, Convolution LSTM will replace the C3D to attain more spatiotemporal relationships in the model. Secondly, the more effective generative model-GAN will be added to better learn a generative model and involved in representation learning. Thirdly, in fact, we are just doing the foundation usage of the frontier predictive learning[49]. Next time we will start from this experiment and focus more on the accessible as well as philosophic aspect of our predictive learning model.

\end{document}